\documentclass[letterpaper, 10 pt, conference]{ieeeconf}
\IEEEoverridecommandlockouts
\usepackage{cite}
\usepackage{amsmath,amssymb,amsfonts}
\usepackage{algorithmic}
\usepackage{graphicx}
\usepackage{textcomp}
\usepackage{xcolor}
\usepackage[ruled,vlined]{algorithm2e}
\usepackage{tikz}
\usepackage{placeins}
\usepackage{balance}
\usepackage{eso-pic}
\usepackage{hyperref}

\newcommand\copyrighttext{%
  \footnotesize \textcopyright 2020 IEEE. Personal use of this material is permitted.
  Permission from IEEE must be obtained for all other uses, in any current or future
  media, including reprinting/republishing this material for advertising or promotional
  purposes, creating new collective works, for resale or redistribution to servers or
  lists, or reuse of any copyrighted component of this work in other works.
  DOI:\href{https://doi.org/10.1109/IROS45743.2020.9341596}{10.1109/IROS45743.2020.9341596}}
\newcommand\copyrightnotice{%
\begin{tikzpicture}[remember picture,overlay]
\node[anchor=south,yshift=10pt] at (current page.south) {\fbox{\parbox{\dimexpr\textwidth-\fboxsep-\fboxrule\relax}{\copyrighttext}}};
\end{tikzpicture}%
}

\newcommand{\note}[1]{#1}
\newcommand{\comment}[1]{}

\def\BibTeX{{\rm B\kern-.05em{\sc i\kern-.025em b}\kern-.08em
    T\kern-.1667em\lower.7ex\hbox{E}\kern-.125emX}}

\title{\LARGE \bf
Deep Tactile Experience: Estimating Tactile Sensor Output from Depth Sensor Data
}

\author{Karankumar Patel, Soshi Iba, Nawid Jamali
\thanks{The authors are with Honda Research Institute USA, Inc.
       {\tt\small \{karankumar\_patel, siba, njamali\}@honda-ri.com}}
}

\begin{document}
\maketitle
\copyrightnotice
\thispagestyle{empty}
\pagestyle{empty}

\begin{abstract}
Tactile sensing is inherently contact based. \note{To use tactile data, robots need to make contact with the surface of an object. This is inefficient in applications where an agent needs to make a decision between multiple alternatives that depend the physical properties of the contact location.}\comment{In order to manipulate the object, often, robots need to make multiple contacts with the object to determine a suitable contact surface. This is inefficient and may cause damage to the object.} \note{We propose a method to get tactile data in a non-invasive manner\comment{, that is, without making contact with the object}. The proposed method estimates the output of a tactile sensor from the depth data of the surface of the object based on past experiences.}
\comment{To get tactile data without making contact with the object, we propose a method that, based on past experience, estimates the output of a tactile sensor from the depth data of the surface of the object.} An experience dataset is built by allowing the robot to interact with various objects, collecting tactile data and the corresponding object surface depth data. We use the experience dataset to train a neural network to estimate the tactile output from depth data alone. We use GelSight tactile sensors, an image-based sensor, to generate images that capture detailed surface features at the contact location. We train a network with a dataset containing 578 tactile-image to depth-map correspondences. Given a depth-map of the surface of an object, the network outputs an estimate of the response of the tactile sensor, should it make a contact with the object. We evaluate the method with structural similarity index matrix (SSIM), a similarity metric between two images commonly used in image processing community. We present experimental results that show the proposed method outperforms a baseline that uses random images with statistical significance getting an SSIM score of  0.84~$\pm$~0.0056 and 0.80~$\pm$~0.0036, respectively.

\end{abstract}


\section{Introduction}
Perception of object properties such as texture, roughness and slipperiness require an agent to make contact with the environment. Humans use their tactile sensors to perceive \note{finger-object} contact properties~\cite{tiest2010tactual}. With sufficient experience, however, humans are also able to perceive such properties from visual perception alone~\cite{tanaka2015investigating , yanagisawa2015effects}, albeit, a rough qualitative estimate of a certain physical property~\cite{fleming2014visual}, for example, the object surface appears to be smooth and slippery. Estimates of contact properties of an object such as roughness and slipperiness can help an agent, in advance, to decide how to interact with the object without making contact with the object. 

\note{The proposed method can be applied to any \comment{robotic} problem where knowledge of physical properties of the point of contact is needed when making a decision, for example, a legged robot can use the method to plan where to plant its feet to avoid a slippery surface. Another application is in grasp planning.\comment{. For example, to} To successfully grasp an object, often, robots need to make multiple contacts with the object to determine a suitable contact location for a stable grasp. Grasping objects that involve multiple contact attempts to generate the tactile data is time consuming and inefficient for the robot. The proposed method can be used during planning to make better informed grasps. It can increase the efficiency of dexterous robots and help them gracefully manipulate objects in their environment. Other applications include haptic rendering where it can be used by blind persons to perceive their surroundings~\cite{wacker2016vibrovision}.}

\comment{Giving robots the ability to asses tactile properties of an object without making contact can increase their efficiency and help them to gracefully manipulate objects in their environment. For example, to successfully grasp an object, often, robots need to make multiple contacts with the object to determine a suitable contact location for a stable grasp. Grasping objects that involve multiple contact attempts to generate the tactile data is time consuming and inefficient for the robot. It 
 may also cause damage to the object. }

\begin{figure}[t]
    \centering
    \includegraphics[width=0.5\textwidth]{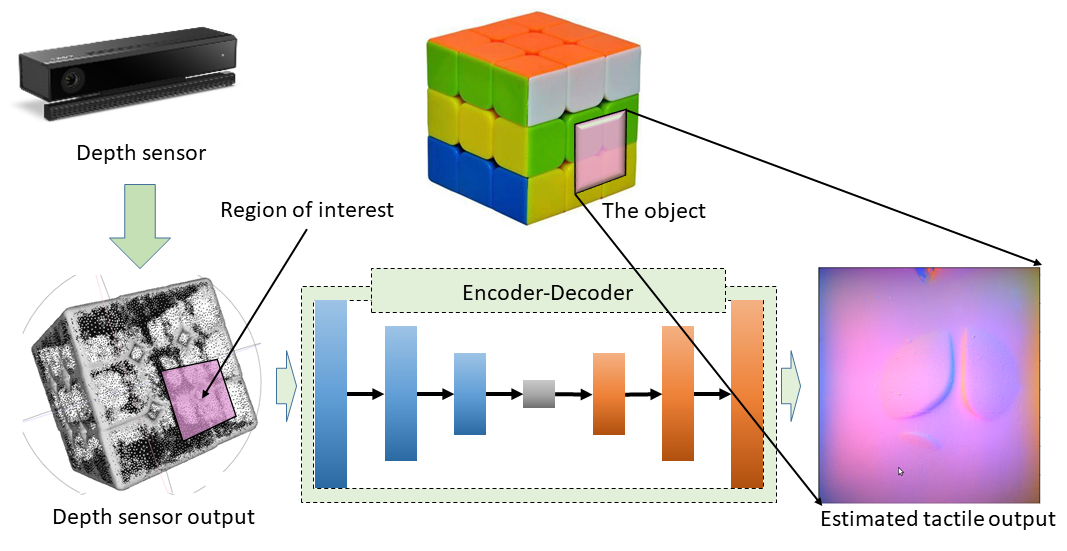}
    \caption[Illustration of the proposed system]{Illustrates a schematic view of the proposed system. The system consists of a depth sensor to sense the structure of the surface of the object. To estimate the tactile sensor output, a region-of-interest, equal to the size of the tactile sensor is selected as the input of an encoder-decoder based neural network. The neural network estimates the output the tactile sensors would have produced if a contact with the object was established.}
    \label{fig:system-overview}
\end{figure}

In this paper we propose a novel idea that uses depth data of the surface of an object to estimate the tactile output, which, \comment{in turn,}can be used by\comment{a robotic system} an agent to estimate physical properties of the object. \comment{Such a system significantly benefits a robot by giving it a non-invasive method to inquire physical contact properties when making decisions. A robot can use the estimate of the tactile sensor output plan a successful grasp or plant its feet on a non-slippery ground in the first attempt.}%
\comment{Such a system significantly benefits a robotic system by eliminating or reducing the number of contacts with an object when making decisions. For example, the robot may be able to grasp an object on the first attempt based on the estimate of tactile sensor output. The proposed method can be applied to any robotic problem where knowledge of physical properties of the point of contact is needed when making a decision, for example, a legged robot can use the method to plan where to plant its feet to avoid a slippery surface.}%
As illustrated in Fig.~\ref{fig:system-overview}, the system consists of a depth sensor to sense the structure of the surface of the object of interest. In order to estimate the tactile output, a region-of-interest\footnote{The region-of-interest can be given by a planner, for example, a grasp synthesizer\cite{bohg2013data} can propose a number of possible \comment{contact}\note{grasp} locations\comment{, for which, it needs}\note{. It can use} tactile sensor estimates to select a better grasp location based on the object's surface features.}, equal to the size of the tactile sensor is selected. The data from the depth sensor in the region-of-interest is used as the input of an encoder-decoder neural network. The neural network estimates the output the tactile sensor would have produced if a contact with the object was established. In other words, the neural network imagines how the tactile sensor will be excited without making any contact with the object. 


\section{Background}

Tactile perception consists of two components: the artificial tactile sensors and the artificial intelligence to interpret the tactile sensor data. Many physical phenomena have been used to create artificial tactile sensors. These include capacitive~\cite{jamali2015new}, peizoresistive~\cite{kerpa2003development}, optical~\cite{hellard2002robust, ohka2004sensing, johnson2009retrographic}, and magnetic~\cite{torres2006soft}. Tactile sensors that have been successfully commercialized include BioTac~\cite{wettels2014multimodal} and GelSight~\cite{johnson2009retrographic} sensors. BioTacs are inspired by human mechanoreceptors that are capable of sensing vibrations and contact forces. GelSight sensors, on the other hand,  are image based. It uses an elastomer and  a camera to encode the shape of a contact surface in the form of an image.

Early work in tactile perception demonstrated that, in principle, tactile sensor signals can be used to differentiate materials~\cite{tada2004sensing, tanaka2007development}. Machine learning has been used to fuse vision and tactile sensory information to improve the accuracy of 3-D object recognition~\cite{kim2004sensor}, differentiate between objects found commonly in a shopping bag~\cite{taddeucci1997approach}, detect slip~\cite{fujimoto2003development, jamali2012slip}, and classify materials~\cite{jamali2011majority, fishel2012bayesian}.

Robotic manipulation is a common task that can benefit from tactile sensing. In order to manipulate an object, the robot needs to plan a grasp configuration that satisfies \note{a set of criteria relevant for the grasping task~\cite{bohg2013data}}, this is referred to as grasp synthesis. Analytical and data-driven approaches have been used to solve the problem of grasp synthesis. Data driven approaches sample grasp candidates and use a metric to rank them~\cite{bohg2013data}. Mahler et al.~\cite{mahler2019learning} propose a network that learns policies to successfully grasp objects in a heap, for a given set of grippers, by training in simulation on synthetic depth data. The authors show that using domain randomization it is possible to transfer the learned policy to a robot. \note{However, the method does not take into account the effect of the contact interaction between the manipulator and the object surface, which may be critical for a successful grasp.} 

Recently, Calandra et al.\cite{calandra2018more} showed that including tactile sensors as additional information can improve over vision only based grasping methods. However, their method requires a contact with the object to generate the tactile sensor data so that the model can predict the stability of the chosen contact location, often, leading to a re-grasp action. The method proposed in this paper, by estimating the tactile output, can eliminate the gripper-object contact required to acquire tactile sensor data. Thereby, making it possible to account for the tactile sensor output without making a contact with the object surface and select a \comment{suitable}\note{better informed} grasp location.\comment{that reduces the need for re-grasping the object.}

Hogan et al.~\cite{hogan2018tactile} propose a tactile-based grasp quality metric. They also propose a re-grasping method that uses the tactile sensor output from the initial contact and use rigid body transformation to simulate tactile sensor output and select a re-grasp location that maximizes the grasp quality metric. The rigid body transformation is not based on the sensed structure of the object, it is a translation of the initial contact tactile data, which may not correspond to the object shape.




A closely related work is presented by Takahashi and Tan~\cite{takahashi2019deep}. The authors propose an encoder-decoder network to learn latent features that map RGB images to tactile data. The mapping is learned by training the network with RGB images of an object as the input and tactile sensor signals of an interaction with the object as the output. After training, the authors propose, use of the latent features to correlate object images to object properties. The authors present experimental results in which tactile data is generated by stroking objects with varying textures and rigidity, and recording an RGB image of the object. The authors show that in the latent space objects are arranged according to friction and rigidity. 

\note{Similar to Takahashi and Tan~\cite{takahashi2019deep}, Li et al.~\cite{li2019connecting} propose an \comment{conditional adversarial networks} encoder-decoder network approach to estimate tactile sensor output. The authors use GelSight tactile sensors, which we also use for our experiments. The input to the network is an RGB image of the entire scene, a reference tactile sensor output and two frames before and after the contact event. The authors present results of a study that used Amazon Mechanical Turk. Human subjects were presented with the ground truth tactile videos and the predicted tactile results along with the vision input. The subjects were asked which tactile video corresponds better to the input vision signal. The authors report that participants identified $46\%$ of the predicted videos to correspond to the input signal for objects in the training set. In case of novel objects, they report $38\%$ of predicted videos were identified to correspond to the input signal.}

The method presented in this paper is different from Takahashi and Tan~\cite{takahashi2019deep} in sensing and the way we interpret the sensor data. We focus on estimating tactile sensor output, which is different from Takahashi and Tan~\cite{takahashi2019deep} who explicitly refrain from inferring the tactile sensor output. Takahashi and Tan~\cite{takahashi2019deep}, and Li et al.~\cite{li2019connecting} use RGB images as input. We use a different sensing modality as the input for our network. We use depth sensor data instead of RGB images. Depth sensor data captures surface structures that directly affect the tactile sensor output.

\section{Methodology}


We present a method that estimates the output of a tactile sensor from depth sensor data of the surface of an object. The intuition behind the method is that a depth sensor captures essential object-surface features that can be used to estimate the output of a tactile sensor. We postulate that it is possible to learn a neural network model that leverages past experience to estimate the tactile sensor output. 
As illustrated in Fig.~\ref{fig:system-overview}, the input to the algorithm is depth sensor data and a region-of-interest. The network outputs an estimate of the tactile sensor output, which, for the tactile sensors used in this paper is an image. Therefore, to train such network we need: a depth-map for the region-of-interest, and the corresponding tactile sensor output. In this section we first give a brief description of the sensors (Section~\ref{sec:sensors}), then explain how we generate the depth-map for a given tactile-object contact (Section~\ref{sec:surface-depth-map}). This is followed by a description of the encoder-decoder network used to model the tactile experience (Section~\ref{sec:network-architecture}).


\subsection{Sensors}
\label{sec:sensors}

We use two Microsoft Kinect-V2 sensors to get depth data on the surface of the object. The tactile data is generated by GelSight~\cite{johnson2009retrographic} tactile sensors. GelSight sensors are optical sensors, unlike other tactile sensors that sense forces and vibrations, a GelSight sensor consists of an elastomer covered with a reflective coating membrane. When the elastomer makes contact with an object surface, the membrane distorts to take on the shape of the surface of the object. A camera is used to record the image of the membrane, which encodes the shape and features of the contact surface. We use these two sensors to generate data for training the neural network.

\begin{figure}[t!]
    \centering
    \includegraphics[width=0.5\textwidth]{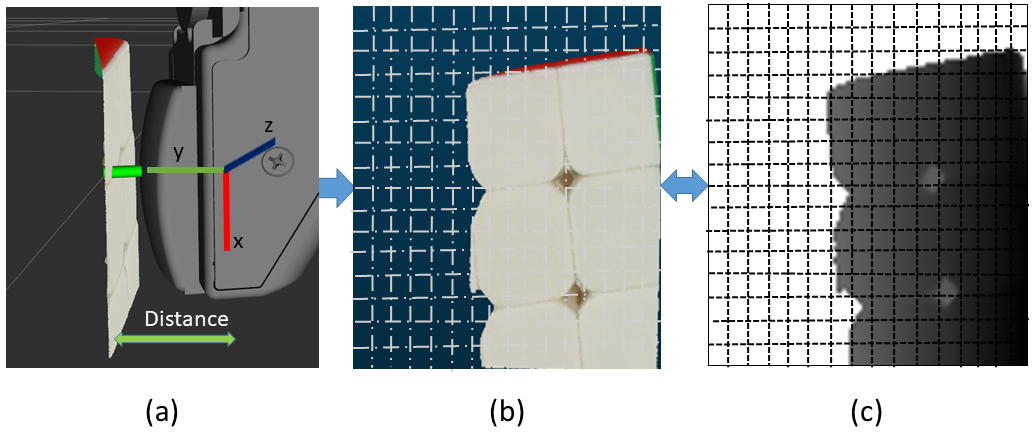}
    \caption[Depth-map generation]{\note{Illustration of depth-map construction.} (a) Object point cloud in the finger's frame of reference. (b) Extracted point cloud of the object surface that is in contact with the tactile sensors. The white lines indicate the bins for depth-map generation. (c) The resulting depth-map after binning. \note{Note: the bins in this example are only for illustration purposes. They do not represent the actual bins used.}}
    \label{fig:depth_map}
\end{figure}

\subsection{Tactile-Object Contact Surface Depth-Map Generation}
\label{sec:surface-depth-map}
We use two Kinect-V2 sensors to get RGB-D data of the environment. Two sensors help to cover a lager area and reduce occlusion. This helps the robot to autonomously sample the object surface. However, two depth sensors are not sufficient to capture the entire object. To ensure that each tactile image has a corresponding complete depth data, we used complete object point clouds provided with the Yale-CMU-Berkeley (YCB) dataset\cite{calli2015benchmarking}. In this section we describe how we build a surface depth-map for each tactile-object contact. \note{Please refer to the accompanied video for a visual illustration of the process.}



\subsubsection{Object Segmentation}
\label{sec:object-segmentation}
The first step is to segment the object of interest. Any algorithm can be used to segment the object. We used Euclidean clustering algorithm~\cite{rusu2010semantic} to segment the object of interest. We crop the point cloud to remove other objects, e.g., objects outside the work space of the robot. We also remove planar surfaces. The filtered point could is used as the input of the Euclidean clustering algorithm, which gives us a single object cluster.


\subsubsection{Registration with Pre-Captured Point Cloud}
\label{sec:registration}

Once we have the object point cloud, then we register a pre-captured point cloud of the object to get a complete point cloud for the object. As described earlier, we use the complete object point clouds that come with the YCB dataset. We used sample consensus alignment algorithm \cite{rusu2009fast} to get an initial coarse alignment. Then, we used the iterative closest point algorithm~\cite{besl1992method} to register the YCB point cloud with the point cloud of the segmented object.


\subsubsection{Object-Tactile Contact Surface Depth-Map}
\label{sec:depth-map-generation}
Once the registration of the complete point cloud of the object described in the previous step is complete, \comment{Once we have \note{registered} the complete point cloud of the object\comment{ registered}, }the robot \note{approaches the object} to make contact with the object at a region-of-interest to collect tactile sensor output. We extract a point cloud patch for the region-of-interest using robot's kinematics. We use this point cloud patch to construct the depth-map. 


Figure~\ref{fig:depth_map} illustrates how a depth-map is generated from the object point cloud. First, the point cloud data is transformed to the tactile sensor's frame of reference with the origin at the center of the tactile sensor. As shown in the Fig.~\ref{fig:depth_map} (a), the $x-z$ plane is parallel to the surface of the tactile sensor, the $y$-axis corresponds to the distance of the object from the surface of the tactile sensor. That is, a point on the surface of the tactile sensor will have a $y$ value of zero, a value of greater than zero means the point is not making contact with the surface of the tactile sensor. Similarly, a point with value less than zero indicates that it has made an indentation in the tactile sensor's elastomer.

The first step is to filter a volume ($x,y,z$) of point cloud that corresponds to the contact area between the object and the tactile sensor. To construct the depth-map, we divide the $x-z$ plane into $m \times k$ bins, indicated by the white dashed lines in Fig.~\ref{fig:depth_map} (b). The value of each bin is set to the minimum $y$ value of points in that bin's $(x~,~z)$ range. Figure~\ref{fig:depth_map}(C) shows a depth-map after binning the cloud patch into $100 \times 100$ bins.

\subsection{Network Architecture}
\label{sec:network-architecture}
\begin{figure}[t]
    \centering
    \includegraphics[width=0.5\textwidth, height=3.8cm]{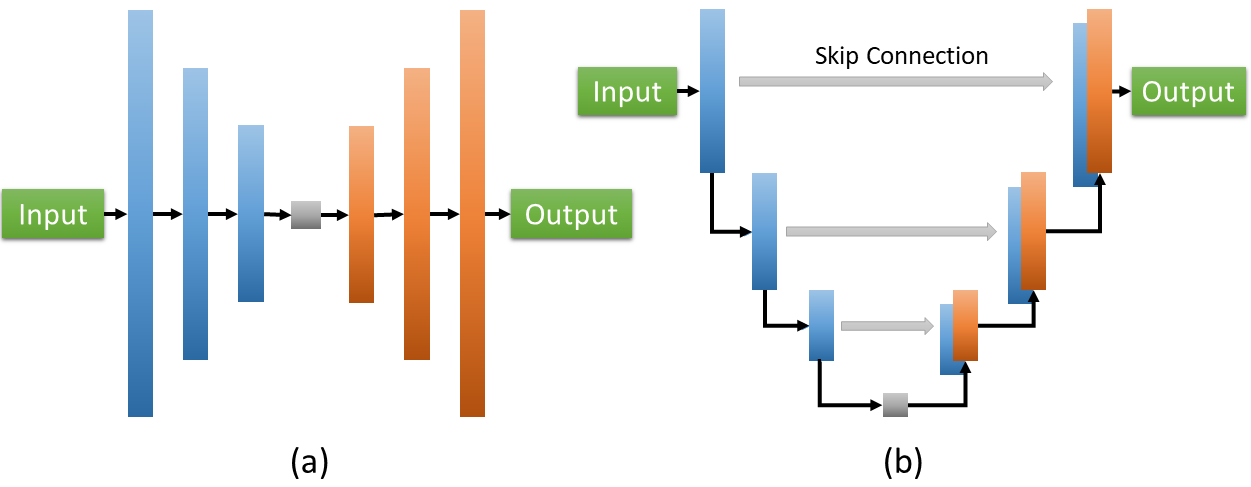}
    \caption[]{Illustrations of encoder-decoder architecture. (a): An encoder-decoder network. (b): A U-Net encoder-decoder network with skip connections. Skip connections help to shuttle low level information of an input from encoder to decoder.}
    \label{fig:u-net}
\end{figure}
We adapt our network architecture from \cite{isola2017image}. We use conditional generative adversarial networks (cGAN) \cite{mirza2014conditional} to model the tactile sensor output estimator. Generative adversarial networks~\cite{goodfellow2014generative} consist of a generator network, $G$, and a discriminator network, $D$. The discriminator tries to learn a loss that can distinguish whether a given input is real or fake. On the other hand, the generator tries to learn how to minimize this loss for fake input. 
Conditional GANs are a form of GANs where the discriminator network is conditioned with an input. So, it can learn conditional generative models. This makes cGANs suitable for our problem, where we condition on an input depth-map and generate the corresponding tactile image. 
The objective of GAN can be described as:

\begin{equation}
\begin{split}
\mathcal{L}_{cGAN}(G,D) &=  {\mathop{\mathbb{E}}}_{x,y}[logD(x,y)] + \\
& {\mathop{\mathbb{E}}}_{x,z}[1-logD(x,logG(x,z))], 
\end{split}
\end{equation}

where $x$, $y$ and $z$ are network input, label and random noise respectively. $G$ tries to minimize this objective against an adversarial $D$ that tries to maximize it, i.e. \begin{equation} 
G^* = arg \mathop{min}_{G}\mathop{max}_{D}L_{cGAN}(G,D) .
\end{equation}

We use L1 loss as described in \cite{isola2017image} rather ran L2 loss because it encourages less blurring.

\begin{equation}
\begin{split}
    \mathcal{L}_{L1}(G)={\mathop{\mathbb{E}}}_{x,y,z}[||y-G(x,z)||_1].
\end{split}
\end{equation}

Hence, our final network objective is:

\begin{equation}
G^*=arg\mathop{min}_{G}\mathop{max}_{D}\mathcal{L}_{cGAN}(G,D)+\lambda \mathcal{L}_{L1}(G).
\end{equation}

where $\lambda$ is the learning rate.


The generator is based on the U-Net~\cite{ronneberger2015u}, an encoder-decoder with skip connections as shown in Figure~\ref{fig:u-net}(b). Using the encoder-decoder based network (Figure~\ref{fig:u-net}(a)) for a generator is well-known. In this type of network, we progressively downsample our input until a bottleneck layer. Later, we try to upsample the data from a bottleneck layer. So, there is a chance of losing information while going from the high dimensional space (input) to a low dimensional space (bottleneck layer). U-Net includes skip connections in between each layer~$i$ and layer~$n-i$, where $n$ is the total number of layers. Skip connection concatenates information from layer~$i$ to layer~$n-i$. This helps to shuttle low level information of an input from encoder to decoder. We use same discriminator as a PatchGAN~\cite{li2016precomputed}. It penalizes structure in terms of patches. This helps the network to learn a structural loss. It outputs whether the given image is real or fake by averaging responses for all the patches. All modules of the generator and discriminator use convolution-BatchNorm-ReLu~\cite{ioffe2015batch}.



\section{Experimental Setup}

\begin{figure}[t]
    \centering
    \includegraphics[width=0.5\textwidth]{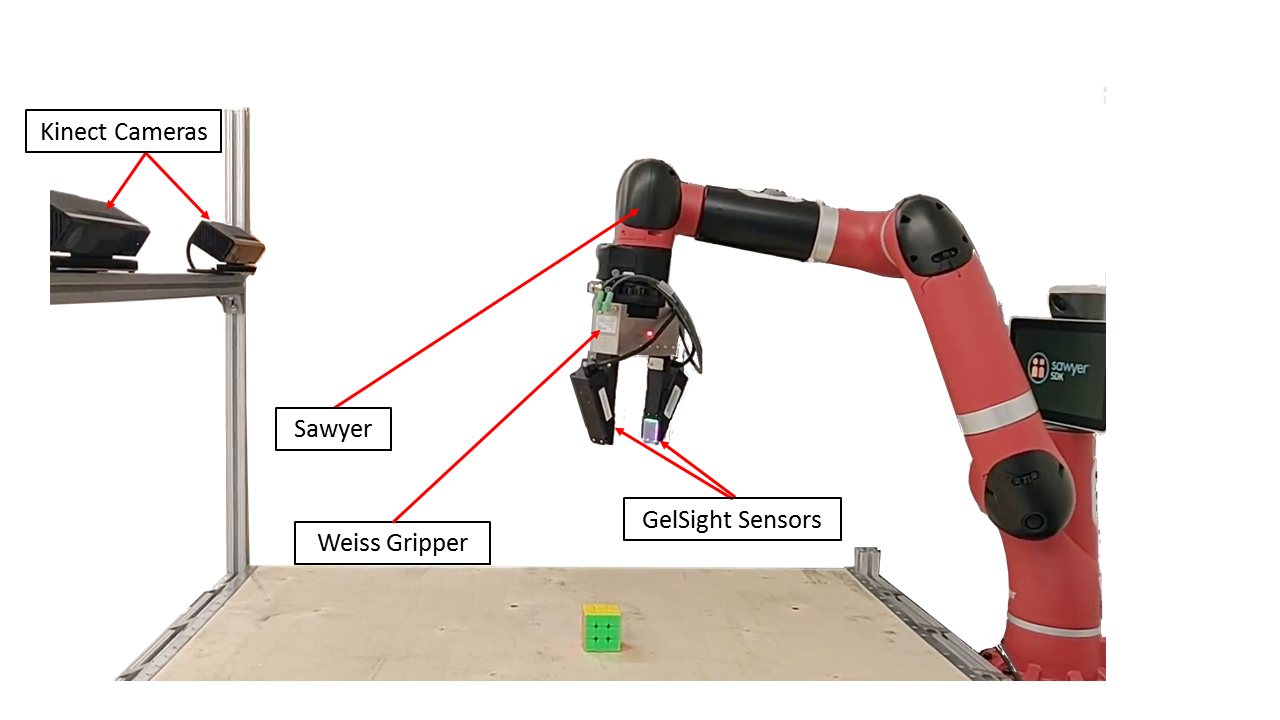}
    \caption[Experimental Setup]{Experimental Setup. It consists of 7-DoF Sawyer robot. The robot is equipped with a Weiss gripper: two-finger parallel jaw gripper. A GelSight sensor is mounted on each finger of the robot. Two Kinect-V2 sensors are used to produce depth data.}
    \label{fig:exp_setup}
\end{figure}

To train and test a model for the proposed tactile sensor estimator we used the experimental setup in Figure~\ref{fig:exp_setup}. It consists of a 7-DoF Sawyer robot by Rethink Robotics. The robot is equipped with a Weiss two-finger parallel jaw gripper. A GelSight~\cite{yuan2017GelSight} sensor is mounted on each finger of the robot. Two Kinect-V2 sensors are used to produce depth data.

\begin{figure}[t]
    \centering
    \includegraphics[width=0.3\textwidth]{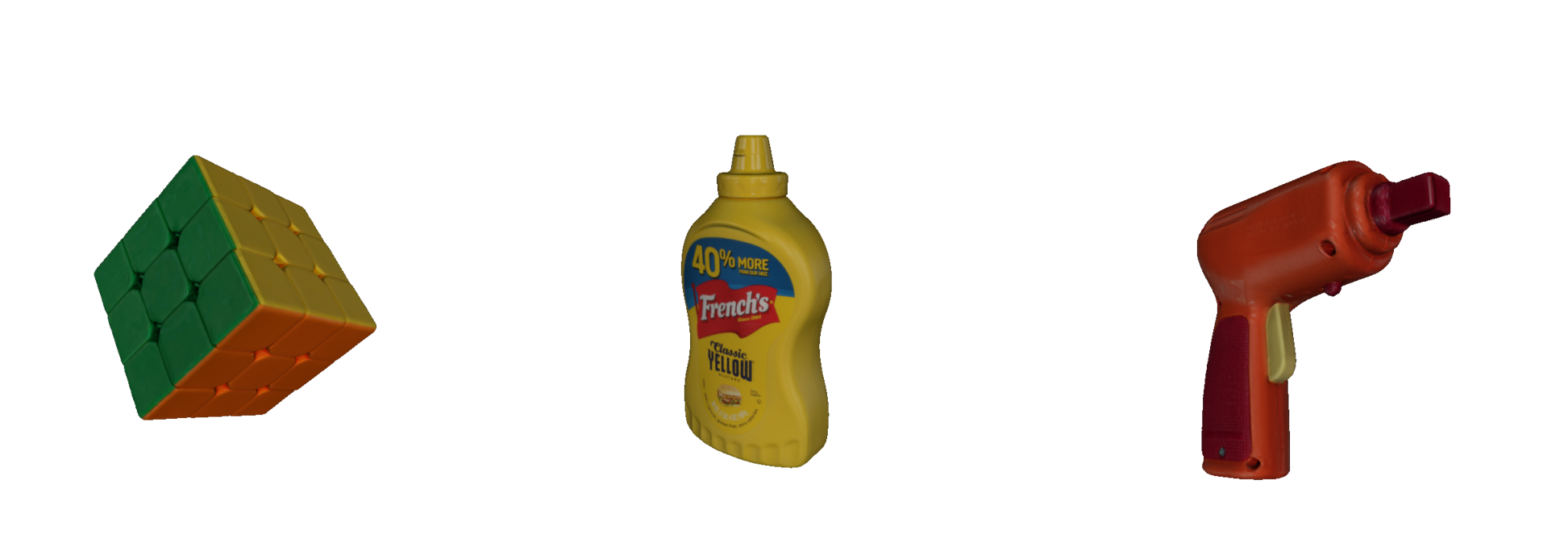}
    \caption[]{Objects used for the data collection.}
    \label{fig:all_objects}
\end{figure}

We used three objects (Fig. \ref{fig:all_objects}) from the YCB Object and Model Set~\cite{calli2015benchmarking} to build the tactile experience dataset. A total of 289 grasps were performed on the objects. With two fingers, resulted in a dataset of 578 (tactile, depth-map) correspondences. The dataset was split into 70\%, 15\% and 15\% for training, validation and testing, respectively.

In the following section we describe the data collection process, then in Section~\ref{sec:training-network} we explain how we train the neural network.

\subsection{Data collection Process}

\begin{figure*}[ht]
    \centering
    \includegraphics[width=\textwidth]{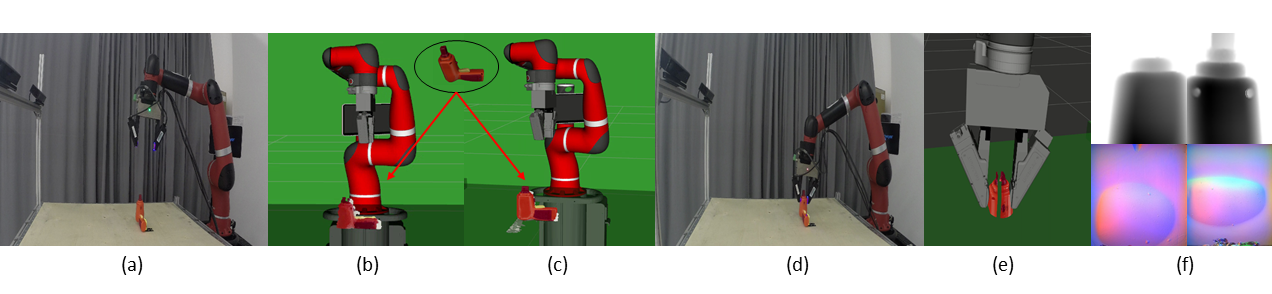}
    \caption[Data collection overview.]{Data collection overview: (a) System setup (b) Object segmentation (c) Registering complete point cloud model of the object (d) Random grasp (e) Extracted point cloud patches (f) Generated depth-maps (top) and tactile sensors output (bottom).}
    \label{fig:data_collection}
\end{figure*}



 \begin{algorithm}[t]
 \label{alg:data-collection}
 \caption{Tactile Experience Data Collection}
 \begin{algorithmic}[1]
 \renewcommand{\algorithmicrequire}{\textbf{Input:}}
 \renewcommand{\algorithmicensure}{\textbf{Output:}}
 \REQUIRE depth data, object model
 \ENSURE  dataset(tactile-image, depth-map)
  \FOR {object in selected objects}
  \STATE object $\leftarrow$ segmentObject(depth)
  \STATE registered-model $\leftarrow$ registerObject(object, model)

   \FOR {grasp in total grasps}
      \STATE $(x, y, z, \theta) \leftarrow$ randomGrasp(object)
      \STATE contact-volume $\leftarrow$ extractPatch(registered-model)
      \STATE depth-map $\leftarrow$ generateDepth(contact-volume)
      \STATE add dataset(tactile-image, depth-map)
   \ENDFOR
  \ENDFOR
 \RETURN dataset 
 \end{algorithmic} 
 \end{algorithm}
 
The data collection process is illustrated in Fig.~\ref{fig:data_collection}. During data collection, the object is rigidly attached to the table. This ensures that the object does not shift when the robot makes contact with it\footnote{This is only required during data collection. At run-time the object can be placed freely.}. Then the robot samples the surface of the object at random locations. 

Algorithm~\ref{alg:data-collection} describes the data collection process. The input to the process is the point cloud from the depth sensors and the object model, that is the complete point cloud  from the YCB dataset. The output is a dataset, D, of (tactile-image, depth-map) couples. That is, for each tactile image from a contact with the surface of the object, the process generates the corresponding depth-map.  

For each object, first, the object point cloud is segmented (described in Section~\ref{sec:object-segmentation}). The next step is to register the object model to the current state of the object (described in Section~\ref{sec:registration}). Once the robot has the registered model, it selects a random location $(x,y,z,\theta)$ to sample the object surface. The $(x,y,z)$ is calculated by adding a random value to the  centroid of the object point cloud. The orientation, $\theta_x, \theta_y$, are fixed such that the palm of the gripper is parallel to the table surface. Only  $\theta_z$, of the gripper is changed. The contact force is set to a constant value. 

When the robot makes contact with the object, a volume, $(x_c,y_c,z_c)$, of the point cloud at the contact location is extracted for each finger. The dimensions of the contact-volume are determined by the size of the surface of the tactile sensor, ($x_t, y_t$), and the depth of the surface elastomer ($z_t)$. We use the robot's kinematics to determine the contact location. In practice, we noticed that inaccuracies in the calibration of the robot with the external cameras resulted in a systematic shift in the point cloud with respect to the calculated contact position. Since the error is systematic, taking a larger patch than the size of the tactile sensor will give a learning algorithm sufficient information to account for the systematic error.

The point cloud in the contact-volume for each finger is converted into a depth-map using the method described in Section~\ref{sec:depth-map-generation}. The tactile experience database is then updated with the recorded GelSight images and the corresponding depth-map. The process is repeated for each object.

\subsection{Training the Network}
\label{sec:training-network}

\begin{figure}
    \centering
    \includegraphics[width=0.5\textwidth]{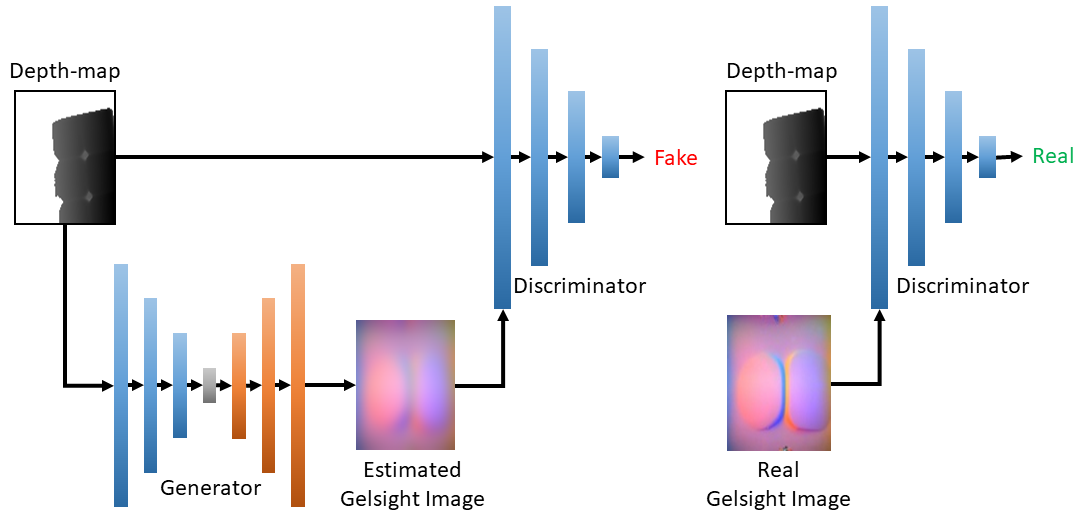}
    \caption[]{Training overview. As shown in figure we provide depth-map as an input to our generator and get estimated tactile image as an output. Discriminator also takes in depth-map and tactile image as an input and outputs whether given tactile image is real or not.}
    \label{fig:training-network}
\end{figure}

As illustrated in Fig.~\ref{fig:training-network}, we use the depth-map and the GelSight images to train the conditional generative adversarial network (cGAN) described in Section~\ref{sec:network-architecture}. 

The input to the network is a $100 \times 100$ image both for the depth-map and the GelSight image. The model is trained using a $70 \times 70$ PatchGAN architecture as the discriminator. We used mini-batch \note{Stochastic Gradient Descent} with Adam solver \cite{kingma2014adam}. We used a learning rate of 0.0002 and momentum parameters \begin{math}\beta_1 = 0.5, \beta_2 = 0.999\end{math}.




\section{Results}

\subsection{Qualitative Results}


\begin{figure}
    \centering
    \includegraphics[height=0.91\textheight]{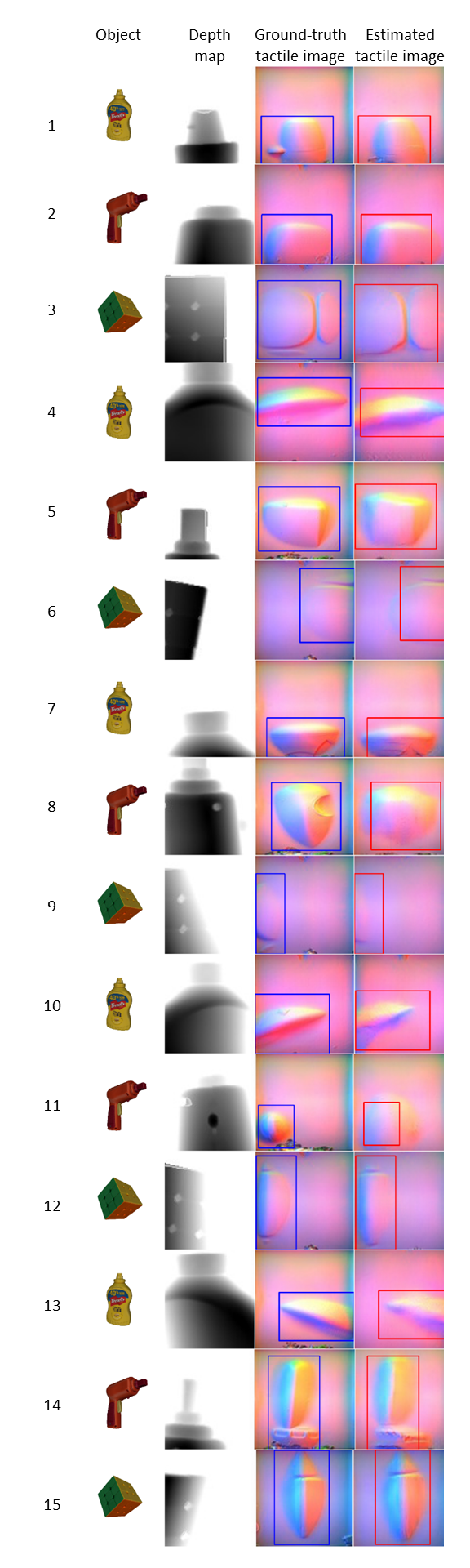}
    \caption[Test results.] {\note{Qualitative results on test data. First column:} object used for data collection. Second column: input to the network -- a depth-map. Third column: ground truth tactile image with a selected template (blue bounding box). Fourth column: estimated tactile sensor output with the template matching result (red bounding box).}
    \label{fig:test_results}
\end{figure}

Figure \ref{fig:test_results} shows the result of estimating a tactile output from depth-maps. In the first column we present a color image of the object to aid in visualizing the surface. The second column is the input to the network, followed by the ground truth tactile image. The last column is the estimate of the network. Visually we can see that the estimated tactile output captures the key features of the object, such as ridges, and surface patterns.

To further analyse the results, we used OpenCV's template matching, which searches an input image for areas that are similar to a given template image. A match is found by sliding the template image across the input image and a similarity metric is calculated. We used the normalized correlation coefficient as the similarity metric. The templates were defined by a human (the authors of this paper) to select a bounding box that captures interesting features \note{(deformed area in GelSight images)} of the ground-truth tactile image. In Fig.~\ref{fig:test_results} the blue bounding boxes show the template and the red bounding boxes show the result of template matching. In some cases, for example, in the ninth example it is not easy to see a correspondence between the ground truth and the estimate. However, in the context of template matching, it is easy to see that the estimate captures important features of the tactile image.



\subsection{Quantitative Results}

\begin{figure}[t]
    \centering
    \includegraphics[width=0.36\columnwidth]{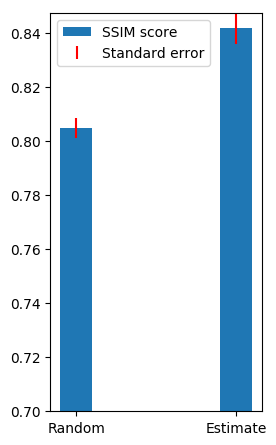}
    \caption[Quantitative results]{Quantitative results. Left bar shows average SSIM score for the baseline. Right bar shows average SSIM score for our method -- between ground truth images and the estimated tactile images.}
    \label{fig:ssim_tm_results}
\end{figure}

Structural similarity matrix (SSIM)~\cite{wang2004image} is a metric to quantify perceived image quality. It is commonly used in image processing to evaluate the quality of processed images. We used SSIM to quantify the quality of the estimated tactile images. It takes two inputs: the ground truth image, and the estimated image. A sliding window approach is used to calculate the similarity metric. For each window in both images, SSIM considers luminance, contrast and structural value. \note{SSIM score ranges between 0 to 1. A score of 1 means two images are perfectly structurally similar while score of 0 means that two images have no structural similarity. A study by Flynn et al.~\cite{flynn2013image} shows that humans can not perceive the difference between a distorted image and a real image at the SSIM score approaches 0.95.} We calculated SSIM scores between ground truth images and the estimated tactile images. We compare the SSIM scores with a baseline SSIM score. 

The baseline score is calculated by taking an average SSIM score between an estimated tactile image and 15 randomly selected ground truth images from the database. That is, for each estimated tactile image, 15 random images were selected and an average SSIM score was calculated. Figure~\ref{fig:ssim_tm_results} shows the result of this analysis. Using our trained model, the average SSIM score between the estimated image and the ground truth is 0.84 $\pm $ 0.0056, compared to the baseline which achieves an average score of 0.80 $\pm$ 0.0036. The results suggest that our model outperforms the baseline with statistical significance. 





\subsection{Effect of Depth Sensor Resolution}

\begin{figure}
    \centering
    \includegraphics[width=0.5\textwidth]{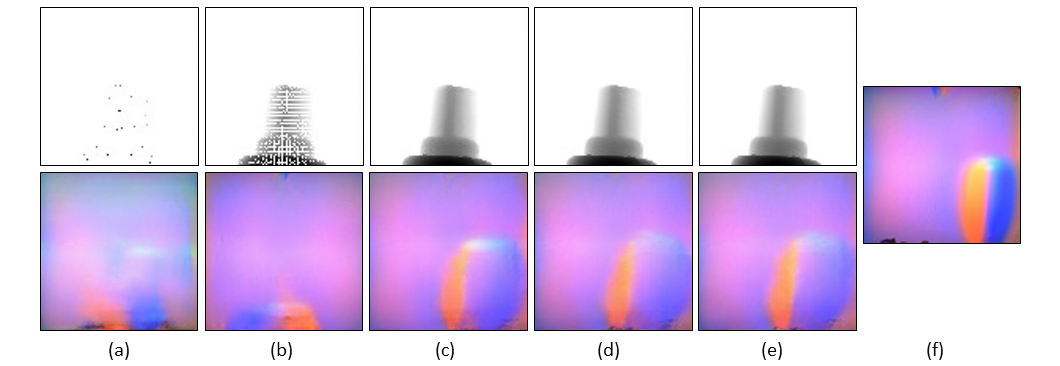}
    \caption[]{Figures a--e show in the top row the depth-maps generated from different point cloud densities. Point cloud densities are 1, 10, 20, 40 and 80 points/cm$^3$ from left to right, respectively. In the bottom row are the corresponding estimates of the GelSight sensor output produced by our method. Figure f shows the ground truth GelSight output.}
    \label{fig:diff-density}
\end{figure}

We also studied the effect of depth sensor resolution on the tactile estimate. We reduced the depth sensor resolution of our point clouds by downsampling the object's point cloud. We used the voxelgrid filter from the point cloud library, which downsamples a point cloud by using a centroid of all points present in a voxel as an approximation. Figure~\ref{fig:diff-density} shows an example of estimated tactile sensor outputs using depth-maps from different point cloud densities. Table~\ref{table:cloud_density_table} shows the SSIM scores for the tactile estimates on point clouds with different densities. \note{The results suggest that a reduction in the point cloud density has negligible adverse effect on the tactile estimate.}


\begin{table}[t!]
\caption{SSIM scores as a function of point cloud density}
\centering
\begin{tabular}{ |p{3.5cm}|p{3cm}|}
 \hline
 Cloud density (points/$cm^3$) & SSIM $\pm$ std error \\
 \hline
 1 & 0.8238~$\pm$~0.0052\\
 10   & 0.8278~$\pm$~0.0057\\
 20  & 0.8478~$\pm$~0.0056\\
 40 & 0.8480~$\pm$~0.0056\\
 80 & 0.8476~$\pm$~0.0057\\
 \hline
\end{tabular}
\label{table:cloud_density_table}
\end{table}


\section{Conclusion and Future work}
We presented a method that estimates the output of a tactile sensor from depth sensor data. The novelty of this work lies in the way we use depth sensor data to estimate tactile sensor output. \note{A neural network based on conditional GANs is trained with a dataset containing couples of contact-surface depth-map and the corresponding tactile sensor output. At run-time, given a depth-map of a region-of-interest, the network produces an estimate of the tactile sensor output should it make contact with the environment at the region-of-interest. Advantage of using depth data is twofold: first, depth sensor data captures surface structures that directly affect the tactile sensor output. Second, depth data is robust to lighting changes compared to RGB images.}

We presented qualitative and quantitative results that suggest the proposed method can estimate tactile sensor output from the depth data. We also presented results of a study that suggests reduction of point cloud density has negligible adverse affect on the quality of the tactile sensor estimates.

\note{An interesting future direction is to study whether inclusion of RGB data and auditory information to complement the depth data will lead to an improved estimate of the tactile sensor output.}

In future, we would like to increase the number of objects. We would also like to study the effectiveness of the tactile estimates for tasks such as classifying objects. Another area that can benefit from the system is grasp synthesis. We would like to explore application of the proposed method to rank grasps such that a robotic system can grasp an object in a single attempt, eliminating the need to sample the surface for a better grasp location. 

\vspace{3cm}
\FloatBarrier
\bibliographystyle{./bibliography/IEEEtran}
\balance
\bibliography{./bibliography/IEEEabrv,./main}

\end{document}